%% file: main.tex
\documentclass[10pt,twocolumn,letterpaper]{article}

\usepackage[algorithms]{wacv}

\input{preamble}

\definecolor{wacvblue}{rgb}{0.21,0.49,0.74}
\usepackage[pagebackref,breaklinks,colorlinks,allcolors=wacvblue]{hyperref}

\title{\texttt{SATOR}: Backpropagation-Free 3D Test-Time Adaptation via Sample-Adaptive Token Routing}

\author{
Inseok Kong$^1$ \qquad
Geunyoung Jung$^2$ \qquad
Jiyoung Jung$^2$\thanks{Corresponding author.} \\
$^1$Department of Geo Informatics, University of Seoul, South Korea \\
$^2$Department of Artificial Intelligence, University of Seoul, South Korea \\
{\tt\small \{kong7963, gyjung975, jyjung\}@uos.ac.kr}
}

\begin{document}
\maketitle

\begin{abstract}
% Real-world systems often encounter new data ovesr time (test data), which leads to experiencing target domain shifts -> Why TTA is needed
% Existing Test-Time Adaptation(TTA) methods tend to apply computationally heavy and memory-intensive backpropagation-based approaches to handle this -> Existing TTA
% Here, ~ -> What we propose
% Our model ~ -> Method intro
% Extensive experiments on popular benchmarks ~ -> Experiment Result

  % The abstract paragraph should be indented \nicefrac{1}{2}~inch (3~picas) on
  % both the left- and right-hand margins. Use 10~point type, with a vertical
  % spacing (leading) of 11~points.  The word \textbf{Abstract} must be centered,
  % bold, and in point size 12. Two line spaces precede the abstract. The abstract
  % must be limited to one paragraph.
Test-time adaptation (TTA) mitigates the performance degradation of source-pretrained models on unlabeled test data under distribution shifts. Recent backpropagation-free token-purging methods address 3D point cloud distribution shifts by filtering unreliable tokens. However, their reliance on predefined purge-size candidates requires multiple forward passes per sample and increases inference cost. We propose \texttt{SATOR} (Sample-Adaptive TOken Routing), a source-free and backpropagation-free 3D TTA framework that eliminates this search by deriving both the partitioning threshold and routing decision directly from each sample. \texttt{SATOR} computes token abnormality scores from complementary attention and geometric relation graphs. It then derives a sample-specific partition into retained and pruned tokens via Otsu-based separability, without evaluating candidate sizes. Keep-ratio-calibrated routing then selects the pruned prediction only when its entropy, penalized by the keep ratio, is lower than that of the original prediction. Experiments on ModelNet40-C, ScanObjectNN-C, and ShapeNet-C demonstrate that \texttt{SATOR} achieves state-of-the-art accuracy on three 3D corruption benchmarks.

\end{abstract}

\section{Introduction}

Deep neural networks have recently achieved remarkable success in 3D point cloud classification, enabling various applications such as autonomous driving, robotics, and AR/VR~\cite{qi2017pointnet,qi2017pointnet++,wang2019dynamic,thomas2019kpconv,yu2022point,zhang2022point}. However, models trained on clean source data often suffer severe performance degradation when the test distribution differs from the training distribution. In 3D point clouds, such distribution shifts can arise from multiple factors, including sensor variations, occlusions, density fluctuations, and background perturbations. Because these shifts are inevitable in practice, they present significant challenges in deploying reliable 3D perception models to real-world environments.

Test-time adaptation (TTA) addresses this challenge by adapting a source-pretrained model to a target domain during inference, relying solely on unlabeled target samples. TTA has demonstrated notable efficacy in mitigating distribution shifts within the 2D image domain through a wide range of approaches, including entropy minimization~\cite{nado2020evaluating,wang2020tent,liang2020we,niu2023towards}, pseudo-labeling~\cite{lee2013pseudo}, graph-based refinement~\cite{boudiaf2022parameter}, augmentation-based aggregation~\cite{zhang2022memo}, classifier and batch normalization adjustments~\cite{iwasawa2021test,mirza2022norm}, and test-time training~\cite{sun2020test}. Nevertheless, transferring these strategies directly to 3D point clouds is non-trivial. Unlike 2D images, 3D point clouds are inherently sparse, unordered, and non-Euclidean. Furthermore, corruptions in 3D point clouds rarely manifest uniformly across the spatial domain. Real-world 3D corruptions often introduce spurious points, remove informative regions, or disrupt local geometric structures, rendering 3D point cloud adaptation fundamentally distinct from that of 2D images.

In 3D point cloud classification, existing TTA methods span test-time training~\cite{mirza2023mate,Bahri2025SMARTPC}, multi-view aggregation~\cite{bahri2025test,kong2026mamvi}, and diffusion-based input restoration~\cite{shim2024cloudfixer,dastmalchi2025test}, but each either ties adaptation to a modified source-training procedure, scales inference cost with the number of generated views, or relies on a source-trained generative model. Although BFTT3D~\cite{wang2024backpropagation} eliminates backpropagation, it still depends on class-level source prototypes at test time. Purge-Gate (PG)~\cite{yazdanpanah2025purge} advances this paradigm by introducing token purging for source-free, backpropagation-free TTA. However, PG measures token divergence against a single global reference and selects the purge-size by searching over manually specified candidates. Consequently, the purge decision is not derived directly from the current sample, and the adaptation process requires multiple forward passes per sample.

Motivated by these limitations, we introduce \texttt{SATOR} (Sample-Adaptive TOken Routing), a source-free and backpropagation-free 3D TTA framework that derives its token partition and routing decision from each test sample. \texttt{SATOR} computes token abnormality scores from attention and geometric relation graphs within the current sample rather than against a stored global reference. These scores are adaptively fused via Otsu-based separability~\cite{otsu1975threshold} and partitioned using sorted-split Otsu thresholding, yielding a sample-specific token partition without evaluating predefined purge-size candidates. A keep-ratio-calibrated routing mechanism then selects between predictions from the original and pruned token sequences, accounting for the keep ratio in the routing criterion. Overall, \texttt{SATOR} obtains its partition from within-sample token relations rather than from a search over predefined candidates, while preserving the deployment advantages of source-free and backpropagation-free TTA.

The main contributions of this work are summarized as follows:
\begin{itemize}
    \item We propose \texttt{SATOR}, a source-free and backpropagation-free
    3D TTA framework that replaces iterative purge-size search with
    sample-specific token partitioning and adaptive routing.

    \item We design an attention--geometry abnormality scoring module that
    computes per-token scores from two complementary relation graphs
    within the current sample, rather than from divergence to a stored
    global reference, and fuses them adaptively via Otsu separability. The
    fused scores directly induce a sample-specific partition threshold,
    so that the pruning level is directly tied to the current sample
    rather than to manually specified purge-sizes.

    \item We introduce keep-ratio-calibrated routing that penalizes
    predictions obtained after more aggressive pruning, selecting the pruned
    prediction only when its keep-ratio-adjusted entropy is lower than that
    of the original prediction.

    \item We demonstrate through extensive experiments that \texttt{SATOR}
    achieves state-of-the-art mean accuracy on the ModelNet40-C,
    ScanObjectNN-C, and ShapeNet-C corruption benchmarks.
\end{itemize}

\begin{figure*}[t]
    \centering
    \includegraphics[width=\textwidth]{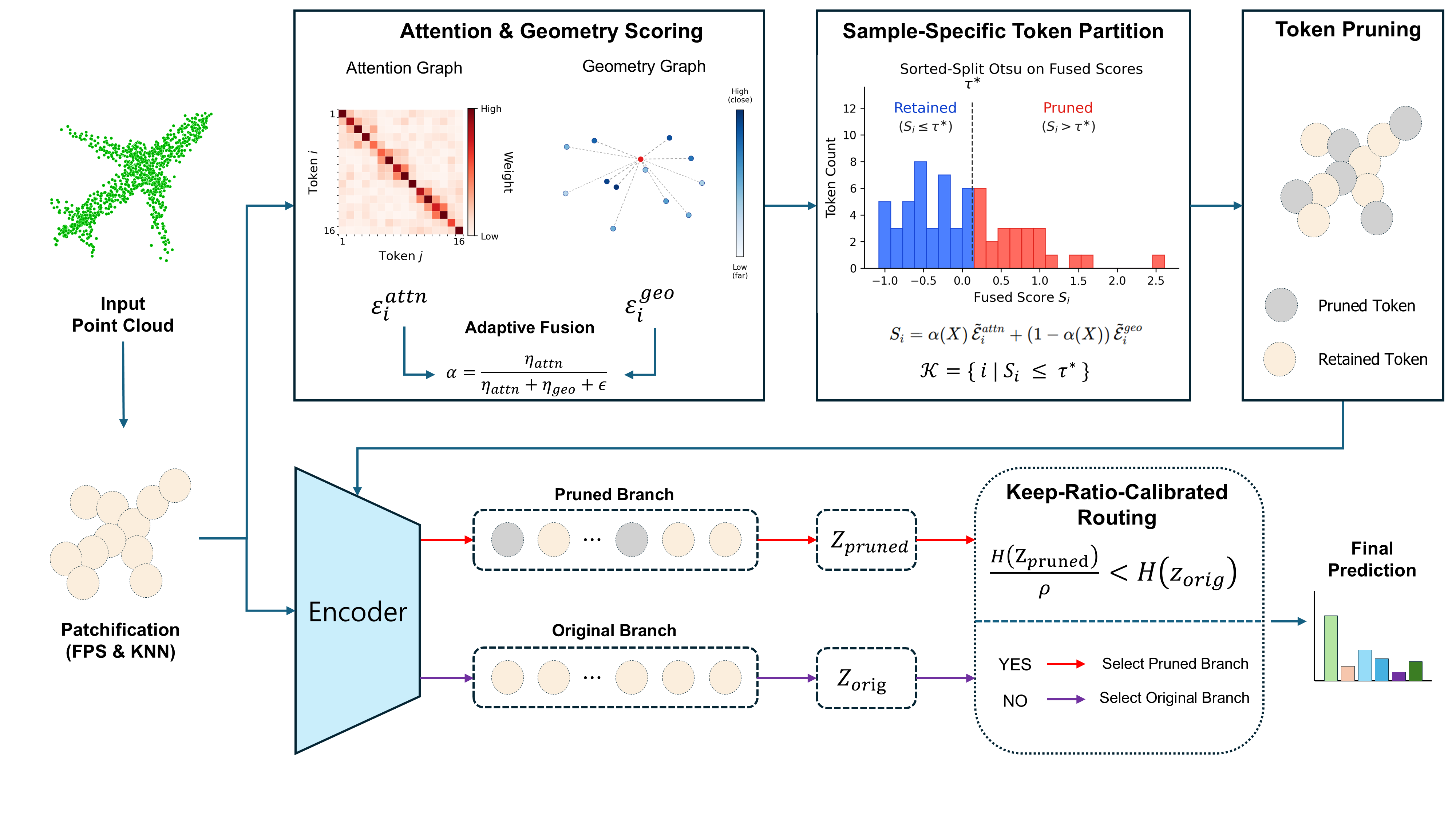}
    \caption{Overall pipeline of \texttt{SATOR}. Token abnormality scores are computed from attention and geometric relation graphs and adaptively fused based on their Otsu separability. A sample-specific token partition is then derived directly from the fused score distribution via sorted-split Otsu thresholding, eliminating iterative purge-size search. A keep-ratio-calibrated routing mechanism selects between the original and pruned token sequence predictions, accounting for the keep ratio in the routing criterion.}
    \label{fig:main}
\end{figure*}

\section{Related Work}

\subsection{Point Cloud Test-Time Adaptation}
Test-time adaptation (TTA) has been widely studied in the 2D image domain through various strategies, including pseudo-labeling, entropy minimization, test-time training, and augmentation-aware prediction aggregation~\cite{lee2013pseudo,sun2020test,wang2020tent,zhang2022memo}. However, extending these ideas to 3D point clouds is non-trivial because point clouds are sparse, unordered, and non-Euclidean. Furthermore, 3D corruptions often manifest irregularly across spatial domains, rendering uniform adaptation strategies less effective.

Existing 3D TTA methods can be categorized by their adaptation mechanisms. TTT-based methods such as MATE~\cite{mirza2023mate} and SMART-PC~\cite{Bahri2025SMARTPC} incorporate auxiliary self-supervised objectives but tie adaptation to a modified source-training procedure, limiting applicability to standard pretrained models. SVWA~\cite{bahri2025test} and MAMVI~\cite{kong2026mamvi} leverage multiple views generated through sampling or masking variations, improving robustness through multi-view aggregation at the cost of higher inference overhead. CloudFixer~\cite{shim2024cloudfixer} and 3DD-TTA~\cite{dastmalchi2025test} rely on iterative diffusion-based input reconstruction and source-trained generative models, making them unsuitable for strictly source-free settings. GSDTTA~\cite{wei20253d} shifts adaptation to the graph spectral domain for efficiency, but still depends on backpropagation. BFTT3D~\cite{wang2024backpropagation} eliminates backpropagation but requires class-level source prototypes at test time. Moving toward efficient source-free adaptation, LN-TTA~\cite{yazdanpanah2026revisiting} and PG~\cite{yazdanpanah2025purge} require neither source data nor gradient updates. LN-TTA recalibrates the affine parameters of layer normalization and leaves the token sequence unchanged. PG proposes two variants: PG-SF (source-free) and PG-SP (source-prototype-based); since our setting is strictly source-free, we compare against PG-SF throughout. However, both PG variants rely on a manual candidate search for purge-size selection, requiring multiple forward passes per sample.

\subsection{Token Scoring and Purging}
Token pruning and merging methods reduce computational cost by identifying and removing redundant tokens~\cite{bolya2022token,rao2021dynamicvit,wu2023ppt,tang2023dynamic}. EViT~\cite{liang2022not} separates tokens into attentive and inattentive groups using CLS attention scores, while ToMe~\cite{bolya2022token} merges similar tokens via bipartite matching. These methods score tokens by their global importance under clean distributions, making them less suited to detecting corruption-induced anomalies that manifest as disruptions in local token interactions.
In 2D TTA, TCA~\cite{wang2025less} shows that token condensation can improve robustness without backpropagation. In 3D TTA, PG~\cite{yazdanpanah2025purge} applies token purging to source-free adaptation, but measures token divergence against a single global reference and selects the purge-size by searching over manually specified candidates, requiring multiple forward passes per sample. \texttt{SATOR} instead scores tokens through attention and geometric relation graphs within each sample, reflecting how each token relates to its local neighbors rather than a stored global reference, enabling search-free partitioning without source statistics.

\subsection{Sample-Adaptive Inference}
Instance-aware inference methods assign each input to a different computational path based on input-specific signals~\cite{wang2018skipnet}. In vision transformers, A-ViT~\cite{yin2022vit} extends this to the token level by learning per-token halting probabilities, enabling sample-specific computation graphs. However, such learned routing policies assume clean distributions and cannot be applied at test time under unknown corruptions.
\texttt{SATOR} replaces learned routing with a parameter-free threshold derived from each sample's token score distribution via sorted-split Otsu thresholding~\cite{otsu1975threshold}, which requires no predefined candidates. The routing decision is further calibrated by the keep ratio, which accounts for the differing token counts of the two branches.

\section{Method}
\subsection{Preliminaries}

\noindent\textbf{Point cloud tokenization.}
A point cloud is an unordered set of 3D points $X=\{p_1,p_2,\dots,p_{N_p}\}\subset\mathbb{R}^3$. Following common point cloud transformer pipelines~\cite{pang2022masked}, we tokenize $X$ into $M$ local patches using furthest point sampling (FPS) and $k$-nearest neighbors (KNN). For the $t$-th token, the patch center is selected as
\begin{equation}
c_t = \arg\max_{p \in X} \min_{c' \in \{c_1,\dots,c_{t-1}\}} \|p-c'\|_2,
\end{equation}
and its neighborhood is defined by
\begin{equation}
\mathcal{N}_t = \{p \in X \mid p \in \mathrm{KNN}(c_t)\}.
\end{equation}
The resulting token is written as $x_t=(c_t,\mathcal{N}_t)$, yielding a token sequence $T=\{x_t\}_{t=1}^{M}$. This representation preserves local geometric structure while enabling transformer-style interactions across patches.

\noindent\textbf{Source-free and backpropagation-free TTA setting.}
We consider a labeled source domain $\mathcal{D}_S=\{(X_i^S,y_i)\}_{i=1}^{N_S}$ and an unlabeled target domain $\mathcal{D}_T=\{X_j^T\}_{j=1}^{N_T}$, with $P_S(X)\neq P_T(X)$. The goal is to improve prediction reliability on target samples without access to source data or backpropagation-based updates at test time.

\subsection{Token Relation Disruption under Distribution Shift}

Distribution shifts in 3D point clouds---such as noise, density changes, and occlusions---alter local patch-token features at the earliest transformer block. Prior token-purging work has analyzed how noisy tokens degrade the attention mechanism under distribution shift, using normalization regularity and high-dimensional random-direction arguments~\cite{yazdanpanah2025purge}. Our analysis is complementary but takes a different perspective: rather than relying on a noise-dominant uniform-attention view or defining token divergence relative to a global reference, we focus on how distribution-induced patch-token feature deviations perturb early token relations. This motivates relation-based abnormality scoring from attention and geometric graphs, which is the basis of Sections~\ref{sec:scoring}--\ref{sec:partitioning}.

Concretely, let $z_i^{S} \in \mathbb{R}^{d}$ denote an ideal clean-domain counterpart of the Block-0 patch-token feature, and write the observed target-domain feature as $z_i^{T}=z_i^{S}+\delta_i$, where $\delta_i$ denotes the feature-space deviation induced by the distribution shift. This notation does not assume a particular corruption model but represents the residual change observed at the patch-token feature level.

In a transformer attention block, pairwise token relations are determined by query-key similarities. For token pair $(i,j)$, the attention logit is
\begin{equation}
s_{ij} =
\frac{\bigl(\mathrm{LN}(z_i+\mathbf{e}_i)W_Q\bigr)^\top
\bigl(\mathrm{LN}(z_j+\mathbf{e}_j)W_K\bigr)}
{\sqrt{d_h}},
\end{equation}
where $\mathbf{e}_i$ and $\mathbf{e}_j$ are positional embeddings and $d_h$ is the head dimension. Even with frozen model parameters, patch-token feature perturbations $\delta_i$ can alter these attention logits and thus change early token relations. The following simple bound, included only to clarify the scoring motivation, formalizes this relation-perturbation view.

\begin{proposition}[Attention-logit sensitivity to patch-token feature perturbations]
\label{prop:logit_perturbation}
Suppose that (i) layer normalization is Lipschitz with constant $L_{\mathrm{LN}}$ on a compact domain where the token-wise standard deviation is bounded away from zero, (ii) the query and key projection matrices satisfy $\|W_Q\|_2 \le B_Q$ and $\|W_K\|_2 \le B_K$, and (iii) the backbone is frozen at test time. Then for any token pair $(i,j)$,
\begin{equation}
\begin{aligned}
|s_{ij}^{T}-s_{ij}^{S}|
&\le
\frac{L_{\mathrm{LN}}}{\sqrt{d_h}}
\Bigl(
  B_K\|q_i^{S}\|_2\|\delta_j\|_2
  +B_Q\|k_j^{S}\|_2\|\delta_i\|_2 \\
&\qquad\qquad
  +L_{\mathrm{LN}}B_QB_K\|\delta_i\|_2\|\delta_j\|_2
\Bigr),
\end{aligned}
\end{equation}
where $q_i^{S}$ and $k_j^{S}$ denote the clean-domain query and key vectors:
\[
q_i^{S}=\mathrm{LN}(z_i^{S}+\mathbf{e}_i)W_Q,\quad
k_j^{S}=\mathrm{LN}(z_j^{S}+\mathbf{e}_j)W_K.
\]
\end{proposition}

The proof is provided in Appendix~A.2. Importantly, this bound is a local sensitivity result: it does not claim that attention collapses to a uniform distribution, nor does it require isotropic random-vector assumptions about the perturbations. It only shows that patch-token feature deviations can change pairwise attention relations even without parameter updates. \texttt{SATOR} therefore uses these perturbed early relations as signals for graph-based abnormality scoring, rather than estimating divergence from an explicit prototype or reference as in prior purging methods~\cite{yazdanpanah2025purge}.

Moreover, through repeated attention mixing in deeper transformer blocks, initially localized perturbations can become diffuse across tokens, making token-level abnormality harder to separate at later layers. This motivates extracting token abnormality scores at Block-0, before repeated mixing obscures local token-level deviations. The block-depth ablation in Table~\ref{tab:block_depth_main} supports this choice.

\subsection{Attention--Geometry Token Abnormality Scoring}
\label{sec:scoring}

The previous section motivates viewing early patch-token relations as sensitive to distribution-induced feature deviations. A single relation graph may not capture all such perturbations because attention relations reflect learned token interactions whereas geometric relations encode spatial neighborhood structure. We therefore compute token abnormality scores at Block-0 from two distinct relation graphs.

\noindent\textbf{Attention relation graph.}
Let $A^{(h)} \in \mathbb{R}^{M \times M}$ denote the Block-0 patch-token attention probabilities of head $h$, where $h=1,\dots,N_h$. We aggregate the attention probabilities across heads and symmetrize them to obtain an undirected attention relation graph
\begin{equation}
W^{attn} = \frac{1}{2}\left(\frac{1}{N_h}\sum_{h=1}^{N_h} A^{(h)} + \left(\frac{1}{N_h}\sum_{h=1}^{N_h} A^{(h)}\right)^\top \right).
\end{equation}

\noindent\textbf{Geometric relation graph.}
The geometric relation graph complements the attention relation graph by measuring consistency among spatially neighboring patches. Let $d_{ij}=\|c_i-c_j\|_2^2$ denote the squared Euclidean distance between patch centers. We connect each token to its $K_g$ nearest neighboring patch centers and define the directed geometric affinity
\begin{equation}
\tilde{W}^{geo}_{ij}
=
\begin{cases}
\exp\left(-d_{ij}/\sigma_g\right), & j \in \mathcal{N}^{geo}_{K_g}(i),\\
0, & \text{otherwise},
\end{cases}
\end{equation}
where $\mathcal{N}^{geo}_{K_g}(i)$ is the set of $K_g$ nearest patch centers to $c_i$, and $\sigma_g$ is the sample-wise mean $K_g$-NN squared distance used as the bandwidth of the geometric affinity kernel. Throughout this section, $\epsilon=10^{-6}$ denotes a small constant for numerical stability. We then row-normalize and symmetrize the graph
\begin{equation}
\bar{W}^{geo}_{ij}
=
\frac{\tilde{W}^{geo}_{ij}}{\sum_{\ell=1}^{M}\tilde{W}^{geo}_{i\ell}+\epsilon},
\qquad
W^{geo}
=
\frac{1}{2}\left(\bar{W}^{geo}+(\bar{W}^{geo})^\top\right).
\end{equation}

\noindent\textbf{Local graph energy as token abnormality.}
Given a relation graph $W$, we define the local graph energy of token $i$ as
\begin{equation}
\mathcal{E}_i(W) = \frac{1}{2} \sum_{j=1}^{M} W_{ij} \|z_i - z_j\|_2^2,
\end{equation}
where $z_i,z_j \in \mathbb{R}^d$ denote the Block-0 patch-token features. This energy increases when token $i$ differs from strongly connected neighboring tokens, capturing both feature inconsistency and disruptions in the underlying relation graph. Applying this energy to the two relation graphs gives the attention-based and geometry-based token abnormality scores $\mathcal{E}_i^{attn}=\mathcal{E}_i(W^{attn})$ and $\mathcal{E}_i^{geo}=\mathcal{E}_i(W^{geo})$, respectively. A formal decomposition of this graph energy under perturbation is provided in Appendix~A.3.

\noindent\textbf{Score normalization.}
Because the attention and geometric relation graphs are constructed from different relation spaces, their raw abnormality scores are not directly comparable. We therefore normalize each score map independently within the current sample
\begin{equation}
\hat{\mathcal{E}}_i^{attn} = \mathrm{Normalize}(\mathcal{E}_i^{attn}), \quad \hat{\mathcal{E}}_i^{geo} = \mathrm{Normalize}(\mathcal{E}_i^{geo}),
\end{equation}
where $\mathrm{Normalize}(\cdot)$ denotes z-score standardization over the $M$ tokens. The normalized scores are then used for sample-specific fusion and partitioning in the next stage.

\subsection{Sample-Specific Token Partitioning}
\label{sec:partitioning}

Given the attention- and geometry-based score maps, the next step is to determine how they should be fused and how the current sample should be partitioned into retained and pruned tokens. Search-based purging methods select from manually specified purge-sizes, leaving the pruning level only indirectly tied to the current sample. In contrast, \texttt{SATOR} derives both the fusion weight and the partitioning threshold directly from the sample's score distribution.

\noindent\textbf{Otsu-based separability for score-map reliability.}
A useful score map should not only identify high-abnormality tokens but also separate them from low-abnormality tokens. We measure this separation using Otsu-based separability~\cite{otsu1975threshold}. For a score map $q$, we define
\begin{equation}
\mathrm{Sep}(q) = \frac{\max_{\tau} \sigma_B^2(\tau)}{\sigma_{\mathrm{total}}^2(q) + \epsilon},
\end{equation}
where $\sigma_B^2(\tau)$ is the between-group variance induced by threshold $\tau$, and $\sigma_{\mathrm{total}}^2(q)$ is the total variance of the score map.

\noindent\textbf{Adaptive fusion of attention and geometry scores.}
We compute the separability of the two normalized score maps as
\begin{equation}
\eta_{attn} = \mathrm{Sep}(\hat{\mathcal{E}}^{attn}), \qquad
\eta_{geo} = \mathrm{Sep}(\hat{\mathcal{E}}^{geo}),
\end{equation}
and use them to weight each relation graph by how cleanly its own score map splits, so that the fusion does not depend on a globally fixed weight. The sample-adaptive fusion weight is
\begin{equation}
\alpha(X) = \frac{\eta_{attn}}{\eta_{attn} + \eta_{geo} + \epsilon},
\end{equation}
and the fused token abnormality score is
\begin{equation}
S_i = \alpha(X)\hat{\mathcal{E}}_i^{attn} + \bigl(1-\alpha(X)\bigr)\hat{\mathcal{E}}_i^{geo}.
\end{equation}
Thus, the fusion is determined by how separable each score map is for the current sample rather than by a fixed weight.

\noindent\textbf{Sorted-split Otsu token partition.}
After fusion, \texttt{SATOR} applies the same separability criterion to the fused score map $S=\{S_i\}_{i=1}^{M}$ to obtain a sample-specific threshold $\tau^\star$. Rather than approximating the score distribution with a histogram, \texttt{SATOR} sorts the $M$ token scores and directly evaluates all $M-1$ candidate split points. This sorted-split approach avoids additional binning hyperparameters and directly matches the discrete nature of token partitioning.

\begin{table*}[t]
\centering
\scriptsize
\setlength{\tabcolsep}{3.2pt}
\caption{Top-1 classification accuracy (\%) across all 15 corruption types in ModelNet40-C. The highest mean accuracy is highlighted in \textbf{bold} and the second highest is \underline{underlined}.}
\label{tab:modelnet}
\resizebox{\textwidth}{!}{%
\begin{tabular}{llccccccccccccccccc}
\toprule
Method & Venue & UNI & GAUSS & BACKG & IMPUL & UPSAM & RBF & RBF-INV & DEN-DEC & DENS-INC & SHEAR & ROT & CUT & DISTORT & OCCL & LIDAR & MEAN & DIFF \\
\midrule
SOURCE & - & 66.53 & 59.12 & 7.21 & 31.69 & 74.72 & 67.71 & 69.77 & 62.28 & 75.08 & 74.35 & 38.13 & 58.63 & 70.02 & 38.53 & 18.64 & 54.16 & 0.00 \\
\midrule
\multicolumn{19}{c}{\textit{Test-Time Training}} \\
\midrule
MATE~\cite{mirza2023mate} & ICCV'23 & 69.80 & 61.80 & 18.90 & 63.90 & 72.50 & 64.00 & 66.00 & 74.00 & 80.80 & 71.00 & 36.70 & 69.20 & 66.30 & 38.40 & 29.90 & 58.88 & 4.72 \\
SMART-PC~\cite{Bahri2025SMARTPC} & ICML'25 & \best{82.40} & \best{80.10} & 12.00 & 67.10 & \best{84.50} & \second{76.00} & \best{78.60} & 67.30 & 72.90 & 73.30 & 43.90 & 72.60 & 73.50 & 37.40 & 24.80 & 63.09 & 8.93 \\
\midrule
\multicolumn{19}{c}{\textit{Backpropagation Method}} \\
\midrule
PL~\cite{lee2013pseudo} & ICML-W'13 & 73.74 & 63.70 & 10.53 & 55.23 & 62.56 & 67.18 & 68.35 & 74.15 & 78.28 & 69.69 & 52.35 & 72.45 & 65.11 & 43.40 & 37.12 & 59.59 & 5.43 \\
TTT-ROT~\cite{sun2020test} & ICML'20 & 61.30 & 58.30 & 34.50 & 48.90 & 66.70 & 63.60 & 63.90 & 59.80 & 68.60 & 55.20 & 27.30 & 54.60 & 64.00 & 40.00 & 29.10 & 53.05 & -1.11 \\
SHOT~\cite{liang2020we} & ICML'20 & 29.60 & 28.20 & 9.80 & 25.40 & 32.70 & 30.30 & 30.10 & 30.90 & 31.20 & 32.10 & 22.80 & 27.30 & 29.40 & 20.80 & 18.60 & 26.61 & -27.55 \\
T3A~\cite{iwasawa2021test} & NeurIPS'21 & 64.06 & 61.55 & 20.14 & 50.12 & 64.83 & 63.74 & 65.03 & 66.57 & 72.29 & 68.03 & 50.49 & 64.67 & 62.80 & 43.64 & 40.28 & 57.22 & 3.06 \\
TENT~\cite{wang2020tent} & ICLR'21 & 73.78 & 70.14 & 18.48 & 59.68 & 76.14 & 70.58 & 72.08 & 73.22 & 79.25 & 73.95 & 54.25 & 70.99 & 70.50 & 46.07 & 43.80 & 63.53 & 9.37 \\
SAR~\cite{niu2023towards} & ICLR'23 & 71.64 & 66.45 & 7.86 & 54.70 & 64.71 & 67.02 & 66.98 & 75.97 & 79.05 & 70.79 & 51.09 & 71.72 & 66.41 & 43.03 & 37.24 & 59.64 & 5.48 \\
MEMO~\cite{zhang2022memo} & NeurIPS'22 & 67.87 & 57.21 & 4.05 & 28.24 & 65.64 & 64.75 & 65.68 & 76.99 & 83.51 & 69.77 & 38.13 & 74.07 & 66.29 & 35.78 & 29.42 & 55.16 & 1.00 \\
\midrule
\multicolumn{19}{c}{\textit{Backpropagation-Free Method}} \\
\midrule
DUA~\cite{mirza2022norm} & CVPR'22 & 65.00 & 58.50 & 14.70 & 48.50 & 68.80 & 62.80 & 63.20 & 62.10 & 66.20 & 68.80 & 46.20 & 53.80 & 64.70 & 41.20 & 36.50 & 54.73 & 0.57 \\
LAME~\cite{boudiaf2022parameter} & CVPR'22 & 67.63 & 58.55 & 4.09 & 31.81 & 63.86 & 64.83 & 65.76 & 76.34 & 82.09 & 68.19 & 37.40 & 71.35 & 65.28 & 36.18 & 29.13 & 54.83 & 0.67 \\
BFTT3D~\cite{wang2024backpropagation} & CVPR'24 & 66.53 & 59.20 & 7.01 & 31.85 & 65.60 & 67.79 & 69.81 & 77.96 & \best{85.25} & 74.35 & 38.17 & 74.80 & 70.02 & 38.61 & 30.51 & 57.16 & 3.00 \\
PG~\cite{yazdanpanah2025purge} & ICCV'25 & 78.12 & 74.47 & \best{40.88} & 66.69 & 81.73 & 73.58 & 75.81 & 79.09 & 82.37 & 77.23 & \second{58.19} & 75.49 & 73.58 & 47.85 & 45.14 & 68.68 & 14.52 \\
LN-TTA~\cite{yazdanpanah2026revisiting} & WACV'26 & \second{80.92} & \second{76.14} & 28.57 & \best{68.92} & 81.36 & 75.28 & 75.77 & \second{80.71} & 83.67 & \best{79.13} & \best{60.25} & \second{77.11} & \second{75.41} & \best{50.89} & \second{51.50} & \second{69.71} & \second{15.55} \\
\rowcolor{dartgray} \texttt{SATOR} & Ours & 79.79 & 75.97 & \second{34.78} & \second{68.02} & \second{81.78} & \best{76.18} & \second{76.58} & \best{82.27} & \second{84.17} & \second{78.33} & 58.08 & \best{80.28} & \best{75.49} & \second{49.92} & \best{52.27} & \best{70.26} & \best{16.10} \\
\bottomrule
\end{tabular}%
}
\end{table*}

\begin{table*}[t]
\centering
\scriptsize
\setlength{\tabcolsep}{3.2pt}
\caption{Top-1 classification accuracy (\%) across all 15 corruption types in ShapeNet-C. The highest mean accuracy is highlighted in \textbf{bold} and the second highest is \underline{underlined}.}
\label{tab:shapenet}
\resizebox{\textwidth}{!}{%
\begin{tabular}{llccccccccccccccccc}
\toprule
Method & Venue & UNI & GAUSS & BACKG & IMPUL & UPSAM & RBF & RBF-INV & DEN-DEC & DENS-INC & SHEAR & ROT & CUT & DISTORT & OCCL & LIDAR & MEAN & DIFF \\
\midrule
SOURCE & - & 77.34 & 71.78 & 8.64 & 54.51 & 77.90 & 75.50 & 76.06 & 85.25 & 76.42 & 80.48 & 57.08 & \best{85.12} & 76.05 & 10.97 & 7.13 & 61.35 & 0.00 \\
\midrule
\multicolumn{19}{c}{\textit{Test-Time Training}} \\
\midrule
MATE~\cite{mirza2023mate} & ICCV'23 & 77.80 & 74.70 & 4.30 & 66.20 & 78.60 & 76.30 & 75.30 & \best{86.10} & \best{86.60} & 79.20 & 56.10 & 84.10 & 76.10 & \second{12.30} & \best{13.10} & 63.12 & 1.77 \\
SMART-PC~\cite{Bahri2025SMARTPC} & ICML'25 & \best{80.80} & \best{78.90} & 8.90 & 60.40 & \best{81.80} & \best{81.10} & \best{81.70} & 84.80 & 78.40 & \second{80.80} & 63.70 & \second{84.90} & \best{79.80} & 11.50 & 8.80 & \second{64.42} & \second{3.07} \\
\midrule
\multicolumn{19}{c}{\textit{Backpropagation Method}} \\
\midrule
PL~\cite{lee2013pseudo} & ICML-W'13 & 69.70 & 66.35 & 8.24 & 59.37 & 62.91 & 71.11 & 72.40 & 80.50 & 79.18 & 74.18 & 56.92 & 80.43 & 71.05 & 9.07 & 9.98 & 58.09 & -3.26 \\
TTT-ROT~\cite{sun2020test} & ICML'20 & 74.60 & 72.40 & \best{23.10} & 59.90 & 74.90 & 73.80 & 75.00 & 81.40 & 82.00 & 69.20 & 49.10 & 79.90 & 72.70 & \best{14.00} & 12.00 & 60.93 & -0.42 \\
SHOT~\cite{liang2020we} & ICML'20 & 44.80 & 42.50 & 12.10 & 37.60 & 45.00 & 43.70 & 44.20 & 48.40 & 49.40 & 45.00 & 32.60 & 46.30 & 39.10 & 6.20 & 5.90 & 36.19 & -25.16 \\
T3A~\cite{iwasawa2021test} & NeurIPS'21 & 66.67 & 64.25 & 13.02 & 59.77 & 67.02 & 65.14 & 64.89 & 70.40 & 66.67 & 65.98 & 54.96 & 69.43 & 63.79 & 10.07 & 10.59 & 54.18 & -7.17 \\
TENT~\cite{wang2020tent} & ICLR'21 & 68.50 & 65.29 & 7.55 & 55.89 & 67.80 & 65.77 & 66.30 & 76.83 & 68.00 & 69.36 & 51.20 & 76.35 & 64.77 & 10.66 & 10.36 & 54.98 & -6.37 \\
SAR~\cite{niu2023towards} & ICLR'23 & 72.57 & 67.54 & 9.43 & 64.06 & 67.02 & 73.12 & 73.88 & 82.44 & 81.55 & 77.46 & 59.59 & 81.51 & 72.80 & 9.54 & 10.05 & 60.17 & -1.18 \\
MEMO~\cite{zhang2022memo} & NeurIPS'22 & 74.89 & 69.54 & 4.25 & 57.69 & 65.80 & 74.52 & 75.72 & \second{85.32} & \second{84.49} & 79.60 & 55.07 & 84.35 & 76.29 & 9.79 & 7.48 & 60.32 & -1.03 \\
\midrule
\multicolumn{19}{c}{\textit{Backpropagation-Free Method}} \\
\midrule
DUA~\cite{mirza2022norm} & CVPR'22 & 76.10 & 70.10 & 14.30 & 60.90 & 76.20 & 71.60 & 72.90 & 80.00 & 83.80 & 77.10 & 57.50 & 75.00 & 72.10 & 11.90 & \second{12.10} & 60.77 & -0.58 \\
LAME~\cite{boudiaf2022parameter} & CVPR'22 & 72.87 & 66.92 & 4.89 & 58.42 & 64.50 & 73.52 & 74.63 & 84.40 & 83.59 & 78.42 & 53.45 & 83.34 & 75.26 & 10.36 & 8.47 & 59.54 & -1.81 \\
BFTT3D~\cite{wang2024backpropagation} & CVPR'24 & 72.58 & 65.91 & 10.57 & 60.86 & 67.78 & 73.45 & 74.38 & 84.55 & 83.52 & 78.58 & 57.96 & 83.71 & 75.17 & 11.95 & 10.00 & 60.73 & -0.62 \\
PG~\cite{yazdanpanah2025purge} & ICCV'25 & 79.21 & 75.43 & \second{22.46} & 68.38 & 78.95 & 76.78 & 77.12 & 83.52 & 80.73 & 79.24 & 64.91 & 83.55 & 76.08 & 10.15 & 9.55 & 64.40 & 3.06 \\
LN-TTA~\cite{yazdanpanah2026revisiting} & WACV'26 & 79.37 & 76.26 & 14.44 & \second{68.66} & 79.56 & \second{78.01} & \second{78.36} & 84.27 & 82.59 & 79.66 & \second{65.24} & 83.63 & 76.95 & 8.99 & 9.32 & 64.35 & 3.00 \\
\rowcolor{dartgray} \texttt{SATOR} & Ours & \second{79.97} & \second{76.42} & 17.24 & \best{70.33} & \second{79.61} & 77.72 & 77.73 & 84.99 & 82.46 & \best{80.90} & \best{65.59} & 84.24 & \second{77.43} & 9.40 & 9.80 & \best{64.92} & \best{3.57} \\
\bottomrule
\end{tabular}%
}
\end{table*}

\begin{table*}[t]
\centering
\scriptsize
\setlength{\tabcolsep}{3.2pt}
\caption{Top-1 classification accuracy (\%) across all 15 corruption types in ScanObjectNN-C. The highest mean accuracy is highlighted in \textbf{bold} and the second highest is \underline{underlined}.}
\label{tab:scanobj}
\resizebox{\textwidth}{!}{%
\begin{tabular}{llccccccccccccccccc}
\toprule
Method & Venue & UNI & GAUSS & BACKG & IMPUL & UPSAM & RBF & RBF-INV & DEN-DEC & DENS-INC & SHEAR & ROT & CUT & DISTORT & OCCL & LIDAR & MEAN & DIFF \\
\midrule
SOURCE & - & 21.69 & 18.76 & 16.70 & 18.42 & 22.20 & 45.96 & 46.99 & 71.95 & 69.36 & 48.88 & 35.63 & 72.98 & 49.40 & 6.71 & 8.78 & 36.96 & 0.00 \\
\midrule
\multicolumn{19}{c}{\textit{Test-Time Training}} \\
\midrule
MATE~\cite{mirza2023mate} & ICCV'23 & 27.50 & 29.40 & 14.30 & 22.20 & 25.60 & 40.80 & 42.00 & \best{73.70} & 63.20 & 45.10 & 35.30 & \best{73.30} & 45.30 & 7.10 & 9.30 & 36.94 & -0.02 \\
SMART-PC~\cite{Bahri2025SMARTPC} & ICML'25 & 27.50 & 39.10 & 19.30 & 21.50 & 29.80 & 44.20 & 48.90 & 68.30 & 60.20 & 49.40 & 45.40 & 70.10 & 49.10 & 8.40 & \second{12.20} & 39.56 & 2.60 \\
\midrule
\multicolumn{19}{c}{\textit{Backpropagation Method}} \\
\midrule
PL~\cite{lee2013pseudo} & ICML-W'13 & \second{45.44} & 43.37 & \second{30.81} & \second{53.36} & \second{44.75} & \second{60.24} & \best{63.34} & 62.31 & 66.95 & 56.11 & 52.32 & 64.72 & \second{59.04} & \second{10.67} & 11.19 & 48.31 & 11.35 \\
TTT-ROT~\cite{sun2020test} & ICML'20 & 30.20 & 34.10 & 16.20 & 22.60 & 22.60 & 32.40 & 32.10 & 45.50 & 45.00 & 34.50 & 29.30 & 47.80 & 36.20 & 7.10 & 8.10 & 29.58 & -7.38 \\
SHOT~\cite{liang2020we} & ICML'20 & 29.50 & 31.60 & 17.60 & 24.80 & 27.20 & 31.00 & 32.40 & 40.70 & 35.00 & 30.20 & 26.60 & 36.60 & 29.30 & 10.50 & \best{12.40} & 27.69 & -9.27 \\
T3A~\cite{iwasawa2021test} & NeurIPS'21 & 35.11 & 46.82 & 13.08 & 30.64 & 38.55 & 50.26 & 53.18 & 68.50 & 63.51 & 50.60 & 46.64 & 69.02 & 53.53 & 7.23 & 10.16 & 42.46 & 5.49 \\
TENT~\cite{wang2020tent} & ICLR'21 & 35.28 & 49.74 & 14.46 & 29.60 & 38.04 & 50.43 & 51.81 & 71.60 & 68.85 & 53.70 & 44.23 & 71.43 & 55.77 & 7.57 & 9.29 & 43.45 & 6.49 \\
SAR~\cite{niu2023towards} & ICLR'23 & \best{46.99} & 42.00 & \best{30.98} & \best{53.70} & 43.37 & \best{60.59} & \second{61.79} & 62.48 & 67.64 & \best{59.55} & \best{53.01} & 65.58 & \best{61.62} & \best{10.84} & 11.53 & \second{48.78} & \second{11.82} \\
MEMO~\cite{zhang2022memo} & NeurIPS'22 & 18.07 & 16.35 & 20.14 & 22.03 & 16.87 & 54.39 & 58.52 & 58.18 & 61.10 & 55.94 & 42.69 & 58.52 & \second{59.04} & 9.64 & 10.15 & 37.44 & 0.48 \\
\midrule
\multicolumn{19}{c}{\textit{Backpropagation-Free Method}} \\
\midrule
DUA~\cite{mirza2022norm} & CVPR'22 & 30.50 & 40.10 & 10.20 & 23.60 & 29.90 & 43.70 & 46.10 & 68.30 & 66.30 & 48.50 & 38.90 & 68.70 & 48.40 & 8.60 & 8.10 & 38.66 & 1.70 \\
LAME~\cite{boudiaf2022parameter} & CVPR'22 & 18.07 & 16.35 & 21.17 & 21.86 & 17.21 & 53.53 & 57.83 & 57.49 & 61.62 & 53.53 & 41.82 & 56.63 & 56.28 & 9.98 & 9.81 & 36.88 & -0.08 \\
BFTT3D~\cite{wang2024backpropagation} & CVPR'24 & 16.35 & 14.97 & 14.29 & 17.04 & 13.43 & 38.90 & 42.00 & 69.71 & 70.74 & 41.14 & 28.92 & 69.19 & 43.20 & 6.54 & 8.61 & 33.00 & -3.96 \\
PG~\cite{yazdanpanah2025purge} & ICCV'25 & 39.59 & 52.67 & 20.83 & 44.58 & 40.62 & 53.70 & 56.11 & 68.85 & 72.12 & 54.56 & 48.88 & 70.91 & \second{59.04} & 10.16 & 9.12 & 46.78 & 9.82 \\
LN-TTA~\cite{yazdanpanah2026revisiting} & WACV'26 & 43.55 & \best{58.52} & 25.30 & 46.47 & 43.89 & 56.45 & 57.14 & \second{71.95} & \second{73.32} & 55.08 & 48.71 & 71.95 & 57.14 & 9.12 & 8.95 & 48.50 & 11.54 \\
\rowcolor{dartgray} \texttt{SATOR} & Ours & 41.32 & \second{57.47} & 24.31 & 50.52 & \best{45.31} & 56.42 & 59.20 & 70.31 & \best{76.39} & \second{56.42} & \second{52.60} & \second{73.26} & 58.16 & 9.20 & 10.59 & \best{49.43} & \best{12.47} \\
\bottomrule
\end{tabular}%
}
\end{table*}

\noindent\textbf{Retained and pruned token sequences.}
Tokens with $S_i \le \tau^\star$ form the retained set $\mathcal{K}$, while tokens with $S_i > \tau^\star$ form the pruned set. This directly determines both which tokens are removed and how many tokens are removed for the current sample, eliminating the need for iterative purge-size search. Unlike PG, which evaluates predefined purge-size candidates through multiple forward passes, \texttt{SATOR} performs search-free sample-specific partitioning by deriving the partition from the sample's own score distribution. The retained token sequence is then forwarded through the remaining backbone to produce the pruned-branch prediction used by the routing stage.

\subsection{Keep-Ratio-Calibrated Routing}

After sample-specific token partitioning produces a pruned token sequence, \texttt{SATOR} decides whether to use its prediction or retain the prediction from the original token sequence.

\noindent\textbf{Entropy and the fraction of retained tokens.}
Token removal can suppress corrupted evidence, but it can also discard useful context, so the pruned prediction is not always preferable. \texttt{SATOR} compares the two branches by their prediction entropy, but entropy alone can favor a prediction whose logits are more sharply peaked, regardless of how many tokens produced it. The following lemma states this relation; it is used here as motivation and does not identify token pruning with temperature scaling. The empirical relation between pruning and entropy is examined in Appendix~C.6.

\begin{lemma}[Entropy decreases under logit sharpening]
\label{lem:entropy_sharpening}
Let $z\in\mathbb{R}^{C}$ be a logit vector with at least two distinct entries, and define the temperature-scaled prediction $p_{\lambda,c} = {\exp(\lambda z_c)}/{\sum_{c'}\exp(\lambda z_{c'})}$. Then
\begin{equation}
\frac{d}{d\lambda}H(p_\lambda) = -\lambda\,\mathrm{Var}_{p_\lambda}(z) \le 0 \quad \text{for all } \lambda > 0.
\end{equation}
\end{lemma}

The proof is provided in Appendix~A.4. Entropy alone is therefore not a sufficient criterion when the two branches are computed from different numbers of tokens: the final routing decision must account for both prediction confidence and the fraction of retained tokens.

\noindent\textbf{Keep-ratio calibration.}%
\label{par:keepratio}
We define the keep ratio $\rho = {|\mathcal{K}|}/{M}$ and the softmax entropy $H(z) = - \sum_{c=1}^{C} p_c \log p_c$, where $p_c = \mathrm{softmax}(z)_c$ and $C$ is the number of classes.
Since $\rho \in (0,1]$, dividing the pruned entropy by $\rho$ penalizes predictions obtained after more aggressive pruning. This prevents low entropy from being favored solely because the prediction was made from a severely reduced token set.

\noindent\textbf{Final routing rule.}
Let $z_{\mathrm{orig}}$ and $z_{\mathrm{prune}}$ denote the logit vectors from the original and pruned token sequences, respectively. \texttt{SATOR} selects the pruned prediction only when its keep-ratio-calibrated entropy is lower than the entropy of the original prediction
\begin{equation}
\label{eq:routing}
\frac{H(z_{\mathrm{prune}})}{\rho} < H(z_{\mathrm{orig}}).
\end{equation}
Otherwise, \texttt{SATOR} retains the prediction from the original token sequence. This makes the final decision adaptive to the current pruning outcome rather than treating token removal as a fixed replacement rule.
\section{Experiments}
\label{sec:experiments}

Detailed dataset descriptions, baseline configurations, implementation settings, additional experiments, and supplementary ablations are provided in the Supplementary Material.

\subsection{Main Results}

\noindent\textbf{ModelNet40-C.}
As shown in Table~\ref{tab:modelnet}, \texttt{SATOR} achieves the highest mean accuracy of 70.26\%, the largest gain over the source-only model (+16.10) among all backpropagation-free methods, ahead of LN-TTA (69.71\%) and PG (68.68\%). It exceeds PG on 13 of the 15 corruptions, with the largest margins on LiDAR distortion (+7.13), cutout (+4.79), and occlusion (+2.07). These corruptions involve spatial distortion or token removal, where within-sample scoring is most informative.

\noindent\textbf{ShapeNet-C.}
ShapeNet-C is the benchmark on which adaptation is hardest to obtain: 10 of the 14 compared methods fall below the source-only model. As shown in Table~\ref{tab:shapenet}, \texttt{SATOR} obtains the best mean accuracy of 64.92\% and the largest improvement over the source-only model among all compared methods (+3.57), ahead of PG (+3.06) and LN-TTA (+3.00). It exceeds PG on 13 of the 15 corruptions.

\noindent\textbf{ScanObjectNN-C.}
ScanObjectNN-C is the only benchmark derived from real-world scans, where corruptions compound with natural scan artifacts. As shown in Table~\ref{tab:scanobj}, \texttt{SATOR} achieves the highest mean accuracy of 49.43\%, again the largest gain over the source-only model (+12.47) among backpropagation-free methods, surpassing LN-TTA by 0.93 and SAR, the strongest backpropagation-based baseline, by 0.65. It exceeds PG on 13 of the 15 corruptions, with the largest margins on impulse noise (+5.94) and gaussian noise (+4.80).

\subsection{Ablation and Analysis}

\noindent\textbf{Token-Level Noise Localization.}
\begin{table}[t]
\centering
\begin{minipage}[t]{0.46\linewidth}
\vspace{0pt}
\centering
\scriptsize
\setlength{\tabcolsep}{3pt}
\renewcommand{\arraystretch}{1.5}
\captionof{table}{Token-level noise localization on background corruption (ScanObjectNN-C).}
\label{tab:background_anomaly_auroc}
\begin{tabular}{lcc}
\toprule
Score & AUROC & AP \\
\midrule
Attn     & 0.751 & 0.765 \\
Geo      & 0.794 & 0.683 \\
Adaptive & \best{0.839} & \best{0.783} \\
\bottomrule
\end{tabular}
\end{minipage}\hfill
\begin{minipage}[t]{0.48\linewidth}
\vspace{0pt}
\centering
\scriptsize
\setlength{\tabcolsep}{3pt}
\renewcommand{\arraystretch}{1.00}
\captionof{table}{Runtime and memory on ScanObjectNN-C. FPS: samples/s; Mem.\ in GB.}
\label{tab:efficiency}
\resizebox{\linewidth}{!}{%
\begin{tabular}{lccc}
\toprule
Method & Acc. & FPS & Mem. \\
\midrule
BFTT3D & 33.00 &  55.9 & 8.77 \\
PG        & 46.78 & 174.1 & 1.38 \\
LN-TTA & 48.50 & \best{790.0} & 1.38 \\
\rowcolor{dartgray}
\texttt{SATOR} & \best{49.43} & 483.2 & 1.38 \\
\bottomrule
\end{tabular}%
}
\end{minipage}
\end{table}
We evaluate whether \texttt{SATOR}'s token abnormality scores localize corrupted tokens. We use background corruption in ScanObjectNN-C, where inserted background points provide a token-level proxy for corruption. As shown in Table~\ref{tab:background_anomaly_auroc}, the fused score achieves the highest AUROC, outperforming both the attention-only and geometry-only scores. This suggests that attention and geometric relation graphs capture complementary corruption signals: the geometry score ranks abnormal tokens more reliably overall with higher AUROC, while the attention score localizes corrupted tokens more precisely with higher AP. Their fusion achieves the best performance on both metrics, indicating that the two graphs provide complementary information for token partitioning.

\noindent\textbf{Runtime and Memory Footprint.}
We compare \texttt{SATOR} with representative backpropagation-free 3D TTA methods. As shown in Table~\ref{tab:efficiency}, \texttt{SATOR} achieves the best accuracy--efficiency trade-off among the compared backpropagation-free methods, reaching higher accuracy than PG (49.43\% vs.\ 46.78\%) while improving throughput by 2.8$\times$ (483.2 vs.\ 174.1 FPS). Compared with BFTT3D, \texttt{SATOR} achieves higher accuracy with 8.6$\times$ faster throughput (483.2 vs.\ 55.9 FPS) and substantially lower memory usage (1.38 vs.\ 8.77 GB). Against LN-TTA, \texttt{SATOR} achieves higher mean accuracy on all three benchmarks at the same memory footprint while remaining competitive in throughput, still ahead of PG and BFTT3D.

\begin{figure*}[t]
\centering
\begin{minipage}[t]{0.54\textwidth}
\vspace{0pt}
\centering
\scriptsize
\setlength{\tabcolsep}{3.6pt}
\renewcommand{\arraystretch}{1.04}
\captionof{table}{Scoring block depth ablation on ScanObjectNN-C.}
\label{tab:block_depth_main}
\resizebox{0.82\linewidth}{!}{%
\begin{tabular}{lcccccc}
\toprule
Block & 0 & 1 & 2 & 3 & 6 & 11 \\
\midrule
Acc. & \best{49.43} & 48.13 & \second{48.23} & 48.16 & 47.66 & 47.36 \\
\bottomrule
\end{tabular}%
}

\vspace{10pt}

\centering
\scriptsize
\setlength{\tabcolsep}{3.2pt}
\captionof{table}{Routing calibration sensitivity across three benchmarks.}
\label{tab:routing_calibration}
\resizebox{\linewidth}{!}{%
\begin{tabular}{lccccc}
\toprule
Dataset & Ent. & $\gamma=0.5$ & $\gamma=1$ & $\gamma=1.5$ & $\gamma=2$ \\
\midrule
ModelNet40-C   & 69.69 & 69.85 & \best{70.26} & 70.07 & 70.02 \\
ScanObjectNN-C & 49.34 & 49.32 & \best{49.43} & 49.27 & 49.12 \\
ShapeNet-C     & 64.79 & 64.82 & \best{64.92} & 64.87 & 64.86 \\
\bottomrule
\end{tabular}%
}
\end{minipage}\hfill
\begin{minipage}[t]{0.42\textwidth}
\vspace{0pt}
\centering
\scriptsize
\setlength{\tabcolsep}{3.6pt}
\renewcommand{\arraystretch}{1.35}
\captionof{table}{Fused-score shape diagnostic on ModelNet40-C. Groups from GMM-BIC; see \textit{Modality Classification} in the supplementary. Accuracy and selection rate in \%.}
\label{tab:fused_score_shape_main}
\resizebox{0.95\linewidth}{!}{%
\begin{tabular}{lccc}
\toprule
Metric & Unimodal & Bimodal & Multimodal \\
\midrule
Fraction & 7.51 & 57.28 & 35.21 \\
Pruned & 63.17 & 66.22 & 68.62 \\
\texttt{SATOR} & \best{65.25} & \best{70.04} & \best{72.06} \\
Gain & +2.08 & +3.82 & +3.44 \\
Selected & 37.52 & 40.41 & 42.75 \\
\bottomrule
\end{tabular}%
}
\end{minipage}
\end{figure*}

\begin{figure*}[t]
\centering
\begin{minipage}[t]{0.46\textwidth}
\vspace{0pt}
\centering
\includegraphics[width=\linewidth]{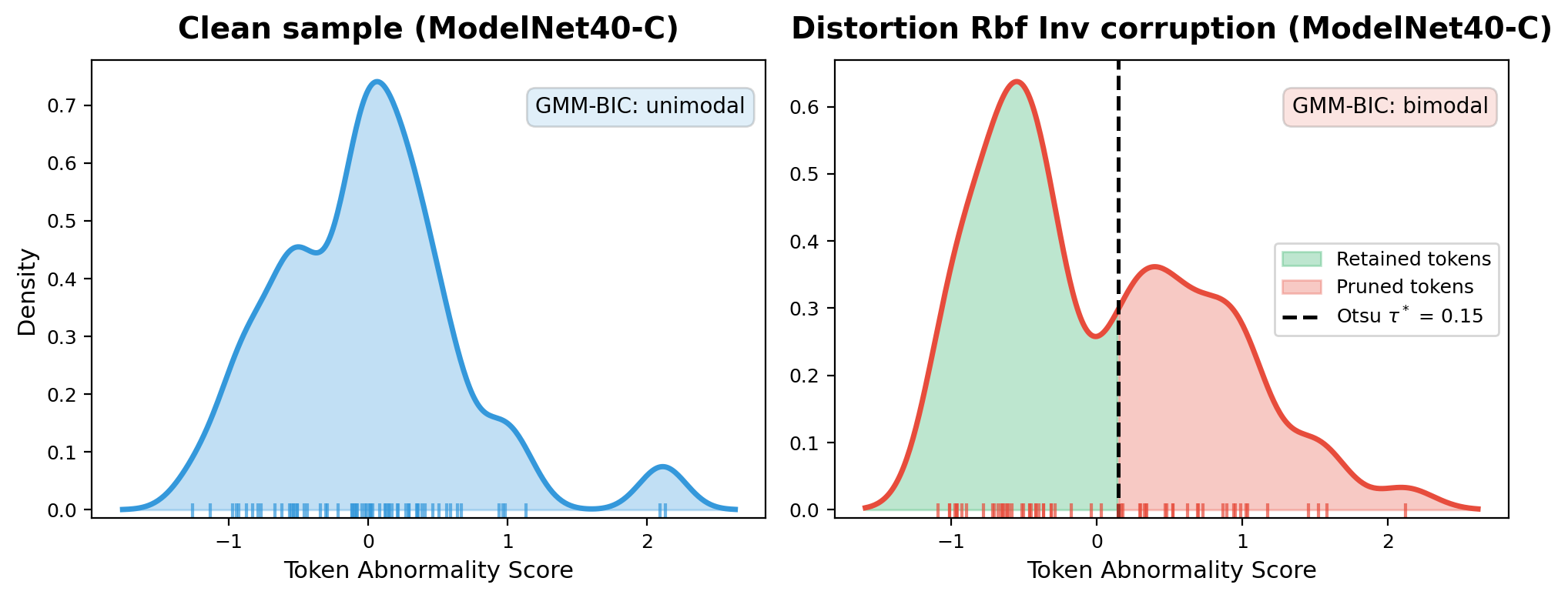}
\captionof{figure}{Fused token abnormality scores on ModelNet40-C. The clean example (left) is unimodal under GMM-BIC; the corrupted example (right) is bimodal, with Otsu threshold $\tau^\star$.}
\label{fig:score_distribution}
\end{minipage}\hfill
\begin{minipage}[t]{0.50\textwidth}
\vspace{0pt}
\centering
\scriptsize
\setlength{\tabcolsep}{3.4pt}
\renewcommand{\arraystretch}{1.4}
\captionof{table}{Component ablation. Final classification accuracy averaged over all corruptions. The w/o Routing columns use the pruned prediction unconditionally and are reported as a diagnostic.}
\label{tab:component_ablation}
\resizebox{\linewidth}{!}{%
\begin{tabular}{l|ccc|ccc}
\toprule
& \multicolumn{3}{c|}{w/o Routing} & \multicolumn{3}{c}{w/ Routing} \\
\cmidrule(lr){2-4}\cmidrule(lr){5-7}
Scoring & MN40-C & ScanObj-C & Shape-C & MN40-C & ScanObj-C & Shape-C \\
\midrule
Geo                & 61.87 & 48.57 & 58.19 & 69.29 & 48.53 & 64.82 \\
Attn               & \best{68.62} & 48.62 & \best{63.96} & 69.84 & 48.37 & 64.69 \\
Fixed $\alpha$=0.5 & 66.48 & 49.02 & 62.17 & 69.85 & 48.77 & 64.79 \\
\rowcolor{dartgray} Adaptive & 66.73 & \best{49.16} & 62.60 & \best{70.26} & \best{49.43} & \best{64.92} \\
\bottomrule
\end{tabular}%
}
\end{minipage}
\end{figure*}

\noindent\textbf{Scoring Block Ablation.}
We examine where token abnormality scoring should be applied within the transformer. Table~\ref{tab:block_depth_main} compares Block-0, 1, 2, 3, 6, and 11 on ScanObjectNN-C with the same routing rule and geometry neighborhood size. Block-0 gives the highest mean accuracy, and on this benchmark accuracy generally decreases as scoring moves to deeper blocks. This suggests that token-level corruption evidence is more separable before extensive token mixing, supporting our choice of Block-0 scoring. Per-corruption-group results for all three benchmarks are provided in Appendix~C.8.

\noindent\textbf{Routing Calibration Sensitivity.}
Table~\ref{tab:routing_calibration} evaluates the sensitivity of the routing criterion to the keep-ratio exponent. Entropy routing compares the two branches without a keep-ratio term. We generalize Eq.~\ref{eq:routing} to $H(z_{\mathrm{prune}})/\rho^{\gamma}$ and vary the exponent $\gamma$; Eq.~\ref{eq:routing} corresponds to $\gamma=1$. Calibrated routing improves over uncalibrated entropy routing on all three benchmarks, by up to 0.57 points on ModelNet40-C. Accuracy peaks at $\gamma=1$ and falls monotonically beyond it on every benchmark, so we use $\gamma=1$ throughout without dataset-specific tuning.

\noindent\textbf{Fused-Score Distribution Shape.}
We group samples by the modality of their fused score distribution, fitting Gaussian mixtures with $k\in\{1,2,3\}$ and selecting $k$ by BIC; 88.8--96.7\% of samples fall in the bimodal or multimodal groups across the three benchmarks (Appendix~C.7). Fig.~\ref{fig:score_distribution} contrasts a clean sample, which is unimodal, with a corrupted sample, which is bimodal with a clear Otsu threshold. Sorted-split Otsu does not assume a bimodal model, and on the unimodal group routing is also the most conservative: as shown in Table~\ref{tab:fused_score_shape_main}, it selects the pruned branch for 37.5\% of unimodal samples against 42.8\% of multimodal ones, and \texttt{SATOR} still improves over the pruned-only branch in every group. Bin-wise analyses over keep ratio and Otsu separability are in Appendix~C.2. Together, these results show that keep-ratio-calibrated routing selectively trusts the pruned branch rather than applying it unconditionally.

\noindent\textbf{Component Ablation.}
Table~\ref{tab:component_ablation} reports final accuracy for each scoring variant. Attention alone, geometry alone, and $\alpha=0.5$ all correspond to a globally fixed weight, and none is safe across datasets: without routing the best variant changes with the benchmark, with a gap of up to 6.75 points on ModelNet40-C. With routing, adaptive fusion is highest on all three, ahead of the fixed-weight variant on every one of them by up to 0.66 points. Both components contribute: on ScanObjectNN-C routing improves only the adaptive score, and replacing our scores with random selection at the same keep ratio loses 1.3--2.4 points on every benchmark. The w/o Routing columns apply pruning unconditionally and isolate the scoring term rather than defining an operating point; under routing the two single scores fall within 0.55 points of each other. Per-corruption results, alternative combination rules, additional controls, and clean-benchmark evaluation are provided in the Supplementary Material.

\section{Conclusion}
\label{sec:conclusion}

We introduced \texttt{SATOR}, a source-free and backpropagation-free framework for 3D test-time adaptation that derives the partitioning threshold and routing decision from each test sample, integrating attention--geometry token abnormality scoring, Otsu-based sample-specific token partitioning, and keep-ratio-calibrated routing. The two scoring signals, attention inconsistency and geometric displacement, prove complementary and separable enough for sample-specific thresholding without a global reference. By deriving the partition from each sample, \texttt{SATOR} removes the repeated evaluations of search-based purging while preserving source-free deployment. Experiments on ModelNet40-C, ScanObjectNN-C, and ShapeNet-C demonstrate that \texttt{SATOR} achieves state-of-the-art accuracy on three 3D corruption benchmarks.

\clearpage
{
    \small
    \bibliographystyle{ieeenat_fullname}
    \bibliography{main}
}

\end{document}

%% file: preamble.tex
\usepackage[utf8]{inputenc}
\usepackage[T1]{fontenc}
\usepackage{nicefrac}
\usepackage{microtype}
\usepackage{colortbl}
\usepackage{capt-of}
\usepackage{pifont}
\usepackage{wrapfig}
\usepackage{amsthm}

\definecolor{dartgray}{gray}{0.88}
\definecolor{oracletext}{gray}{0.55}
\newcommand{\best}[1]{\textbf{#1}}
\newcommand{\second}[1]{\underline{#1}}

\newtheorem{proposition}{Proposition}
\newtheorem{lemma}{Lemma}